\documentclass{lmcs} 


\usepackage{hyperref}
\theoremstyle{plain} 

\usepackage{listings}

\usepackage{amsthm}
\newtheoremstyle{myThm}
{3pt}
{0pt}
{\upshape}
{}
{\bfseries}
{.}
{.5em}
{}
\theoremstyle{myThm}
\newtheorem{myThm}{Theorem}

\newtheoremstyle{myDef}
{3pt}
{0pt}
{\upshape}
{}
{\bfseries}
{.}
{.5em}
{}
\theoremstyle{myDef}
\newtheorem{myDef}{Definition}

\begin{document}

\title[Formalization of Robot Collision Detection Method]{Formalization of Robot Collision Detection Method based on Conformal Geometric Algebra}
\thanks{*Corresponding author(s).}	

\author[Y.~Wu]{Yingjie WU}[a]
\author[G.~Wang]{Guohui Wang*}[a]
\author[S.~Chen]{Shanyan Chen}[b]
\author[Z.~Shi]{Zhiping Shi}[c]
\author[Y.~Guan]{Yong Guan}[d]
\author[X.~Li]{Ximeng Li}[d]

\address{Beijing Engineering Research Center of High Reliable Embedded System, Capital Normal University, Beijing, China}	
\email{yjwu@cun.edu.cn, ghwang@cnu.edu.cn}  

\address{Beijing Key Laboratory of Electronic System Reliability and Prognostics, College of Information Engineering, Capital Normal University, Beijing, China}
\email{sychen@cnu.edu.cn}

\address{International Science and Technology Cooperation Base of Electronic System Reliability and Mathematical Interdisciplinary, Capital Normal University, Beijing, China}
\email{shizp@cnu.edu.cn}

\address{Beijing Advanced Innovation Center for Imaging Theory and Technology, Capital Normal University, Beijing, China}
\email{guanyong@cnu.edu.cn, lixm@cnu.edu.cn}





\begin{abstract}
Cooperative robots can significantly assist people in their productive activities, improving the quality of their works.
Collision detection is vital to ensure the safe and stable operation of cooperative robots in productive activities.
As an advanced geometric language, conformal geometric algebra can simplify the construction of the robot collision model and the calculation of collision distance.
Compared with the formal method based on conformal geometric algebra, the traditional method may have some defects which are difficult to find in the modelling and calculation.
We use the formal method based on conformal geometric algebra to study the collision detection problem of cooperative robots.
This paper builds formal models of geometric primitives and the robot body based on the conformal geometric algebra library in HOL Light.
We analyse the shortest distance between geometric primitives and prove their collision determination conditions.
Based on the above contents, we construct a formal verification framework for the robot collision detection method.
By the end of this paper, we apply the proposed framework to collision detection between two single-arm industrial cooperative robots.
The flexibility and reliability of the proposed framework are verified by constructing a general collision model and a special collision model for two single-arm industrial cooperative robots.
\end{abstract}

\maketitle

\section{Introduction}\label{sec1}

Cooperative robots are widely used to assist humans in heavy tasks \cite {bib1}, which improves work efficiency and serves humans in daily life.
And they are competent for professional work in the fields of industrial production \cite {bib2}, domestic services \cite {bib3}, and medical services \cite {bib4}.
As an example, assembly robot \cite {bib5} can satisfy the requirements of high intensity, high precision and high stability in industrial production.
It is nowadays evident that intelligent behaviours in robots emerge as the interplay among sensory, representation and motor activities in the
real-world, specifically when interacting with humans.
In human-robot interaction, the collision between the two arms of the cooperative robot may injure people or cause serious accidents.
So, it is essential to study the robot collision detection method \cite{bib6}.

There are several methods to solve collision detection \cite{bib7} problems.
The commonly used method to solve the collision detection problem between robots is Euclidean geometry.
Zaheer et al. \cite{bib8} proposed a real-time obstacle avoidance technique for mobile robots based on two-dimensional Cartesian Eigenspace.
Sánchez et al. \cite{bib9} proposed a collision-free navigation algorithm based on bounding box to avoid conflicts between unmanned aerial vehicles (UAVs).
Yun et al. \cite{bib10} proposed a hierarchical collision avoidance algorithm that is robust to kinematic singularity by adding the control input from the geometric safety measure to the conventional repulsive input for collision avoidance.
All of the works above are based on Euclidean geometry.

Mathematical theories in Euclidean geometry have been formally proved \cite{bib11}.
Boutry et al. \cite{bib12} described the formalization of the arithmetization of Euclidean plane geometry in the Coq proof assistant, and introduced Cartesian coordinates to provide characterizations of the main geometric predicates.
Beeson et al. \cite{bib13} used computer proof-checking methods to verify the correctness of their proofs of the propositions in Euclid Book I, then checked them in the well-known and trusted proof checkers HOL Light and Coq.
However, Euclidean geometry relies on the coordinate system \cite{bib14}.
Geometric calculations in Euclidean space, such as Euclidean distance \cite{bib15}, are based on algebraic operations.
These algebraic operations have no geometric meaning if separated from the coordinate system.
So, this kind of method is complex and inflexible.

Compared to Euclidean geometry, conformal geometric algebra (CGA) \cite{bib16} provides a unified algebraic framework for classical geometry.
It frees from the constraints imposed by the coordinate system on the classical geometric language.
CGA is a more general mathematical theory for geometric modelling and a more robust mathematical tool for geometric calculation.
Campos-Macías et al. \cite{bib17} proposed a method to solve the inverse kinematics of a humanoid robot leg with 6 degrees of freedom (6-DOF) using CGA.
Lian et al. \cite{bib18} proposed a geometric error modelling method of parallel manipulators (PMs) based on the visual representation and direct calculation of CGA.
Bayro-corrochano et al. \cite{bib19} addressed the three-dimensional model registration, guided surgery by reformulating screw theory (a generalization of quaternions) and used CGA to solve some critical computational issues in medical robot vision.
Hildenbrand et al. \cite{bib20} realized the industrial application of the forward kinematics algorithm of the serial robot based on the calculation method in CGA.

Generally, CGA-based approaches for geometric modeling \cite{bib21} and geometric calculation have included paper-and-pencil proof \cite{bib22} and Computer Algebra Systems (CAS) \cite{bib23}.
However, paper-and-pencil proof can only handle calculations with a few symbols and will introduce human error.
For accuracy, CAS can perform many symbolic operations using more time and space.
It may have some shortcomings in handling preconditions.
Theorem proving is a highly reliable method that overcomes the limitations of traditional methods.
It can use higher-order logic to prove that the system meets specifications or has the desired properties.
And it can detect minor design errors early, providing an accurate analysis for the robot collision detection method \cite{bib24}.

\begin{figure}[h]
\centering
\includegraphics[scale=0.75]{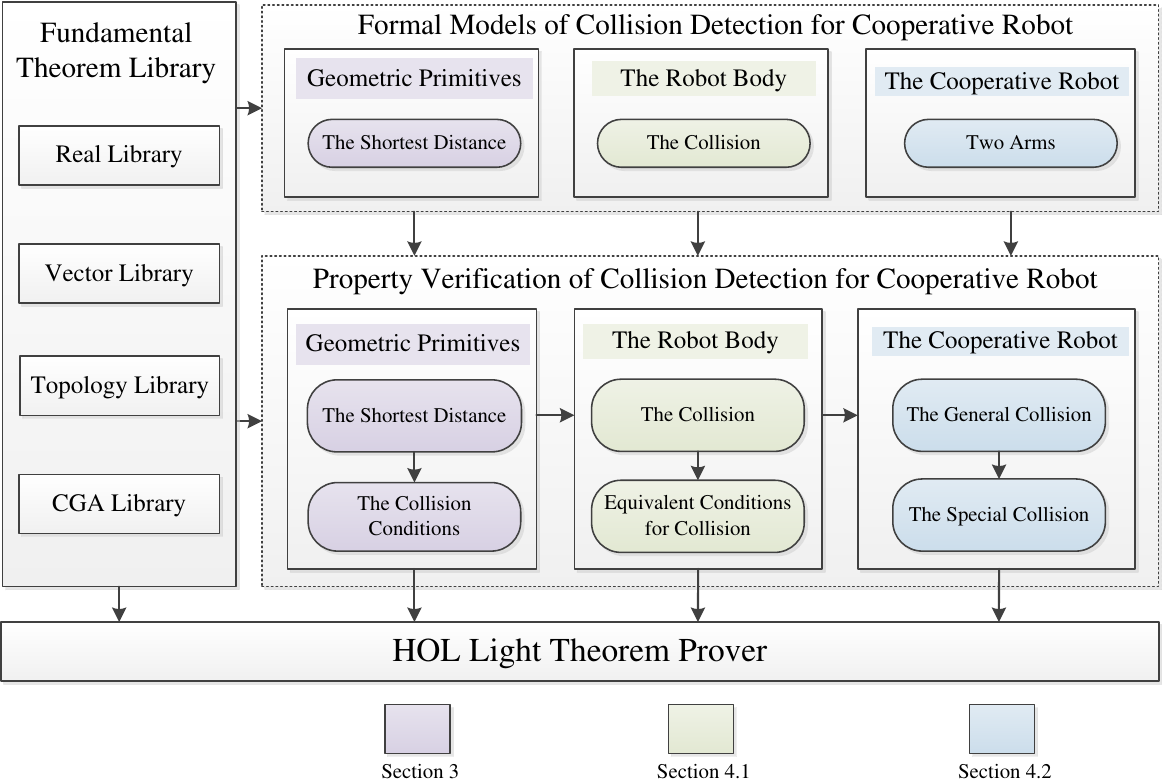}
\caption{The formal verification framework}\label{frame}
\end{figure}

In this paper, we use HOL Light \cite{bib25}, a higher-order logic theorem prover, to propose a formal verification framework for the robot collision detection method.
The proposed framework is also applied to the formal verification of collision detection for two single-arm industrial cooperative
robots.
The formal models and the property verification of collision detection for the cooperative robot are depicted in Fig.\ref{frame}.
The primary motivation for choosing HOL Light is the availability of some fundamental theorem libraries, such as the real library, the vector library, the topology library and the CGA library.
These are some of the foremost requirements for formal verification of robot collision detection.

The major contributions of the paper are given below.
\begin{itemize}
    \item Construction of a formal verification framework of robot collision detection method, which includes the formalization of geometric primitives, the construction of the shortest distance models, and the verification of collision conditions between geometric primitives.
    \item Application of the formal verification framework, which includes the formalization of the robot body, the verification of robot collision equivalence conditions, and the verification of collision detection between two single-arm industrial cooperative robots.
\end{itemize}

The rest of the paper is organized as follows.
Section \ref{sec2} provides an introduction to the HOL Light theorem prover CGA.
Section \ref{sec3} presents the formal verification framework of the robot collision detection method, including the higher-order logic expressions of geometric primitives, the formal models of the shortest distances between geometric primitives and the collision conditions between geometric primitives.
Section \ref{sec4} presents the higher-order logic representation of the robot body, the construction of the collision models and
the application of the robot formal verification framework in two single-arm industrial cooperative robots.
Section \ref{sec5} concludes the paper.

\section{Preliminaries}\label{sec2}
\subsection{HOL Light}\label{subsec21}

HOL Light is an interactive theorem-proving system for conducting mathematical proofs in higher-order logic.
It is highly reusable and reliable.
The theories in HOL Light consist of types, constants, definitions, axioms and theorems.
A new theorem must be verified on the basis of the original inference rules and the basic axioms, or on the theorems which have been verified in HOL Light.
Then, the new theorem is added to the theorem-proof library and continues to serve as the theoretical basis for other new theorems.
There are efficient proof tactics and mathematical theorem libraries in HOL Light, which provide the basis for our work.

The proof scripts of HOL Light are readable and use easy-to-understand logical symbols, functions and predicates to describe definitions and theorems.
Table~\ref{tab1} shows some standard symbols and HOL Light symbols, which are commonly used in this paper.

\begin{table}[h]
\begin{center}
\begin{minipage}{400pt}
\caption{HOL Light symbols and Standard symbols}\label{tab1}%
\begin{tabular}{@{}|p{3.2cm}|p{3.4cm}|l|@{}|}
\hline
HOL Light symbols & Standard symbols & Symbolic meanings \\
\hline
$\texttt {!}$    & $\forall$ & For all  \\
$\texttt {?}$    & $\exists$   & Exists  \\
$\sim$    & $\neg$   & Logical negation  \\
$/\backslash$    & $\wedge$   & Logical and  \\
$\backslash /$    & $\vee$   & Logical or  \\
$\texttt {==>}$    & $\Rightarrow$   & Implication  \\
$\texttt {<=>}$    & $\Leftrightarrow$   & Equivalence  \\
$\texttt {x IN S}$   & x $\in$ S   & x belongs to S  \\
$\texttt {S UNION T}$    & S $\cup$ T   & Union of S and T  \\
$\texttt {S INTER T}$    & S $\cap$ T   & Intersection of S and T  \\
$\texttt {DISJOINT S T}$    & S $\cap$ T = $\emptyset$   & S and T are mutually exclusive  \\
$\texttt {norm(x - y)}$    & $\|$x - y$\|$   & Distance between x and y  \\
$\texttt {--b}$    & -b   & Negative number of scalar b  \\
$\texttt {inv b}$    & $\frac{1}{b}$   & Inverse number of scalar b  \\
$\texttt {\&c}$    & c   & Typecasting from natural number to real  \\
$\texttt {s \% x}$   & s * x   & Scalar multiplication of vector x  \\
$\texttt {s * t}$   & s * t   & Scalar multiplication of scalar t  \\
$\texttt {x dot y}$   & x $\cdot$ y   & Dot product in Euclidean space \\
$\texttt {p inner q}$   & p $\cdot$ q   & Inner product in conformal space \\
{$\texttt {(m..n)}$}    & $\{$ s $\mid$ m $\le$ s $\le$ n$ \} $   & Natural number segment \\
\hline
\end{tabular}
\end{minipage}
\end{center}
\end{table}

\subsection{Conformal Geometric Algebra}\label{subsec22}
As an essential branch of geometric algebra \cite{bib26}, CGA \cite{bib27} is a tool for geometric representation and calculation.
CGA provides a unified algebraic framework for classical geometries.
It shows excellent advantages in establishing geometric models and analyzing geometric data without the limitation of coordinates.
Based on an orthogonal vector basis $\left\{ {{e_{1}},{e_{2}},{e_{3}}} \right\}$ of three-dimensional Euclidean space, CGA introduces two other base vectors, ${e_{+}}$ and ${e_{-}}$.
They are used to define two null vectors, ${e_0}$ and ${e_\infty }$.
The relationships among them are shown as Eq. \eqref{equ1}.
\begin{equation}\label{equ1}
{{\rm{e}}_0} = \frac{1}{2}({e_ - } - {e_ + }),{e_\infty } = {e_ - } + {e_ + }{\rm{, }}{{\rm{e}}_0}^2 = {e_\infty }^2,{e_0}{e_\infty } =  - 1
\end{equation}
An orthogonal vector basis $\{ {e_1},{e_2},{e_3},{e_0},{e_\infty }\}$ are constructed in conformal space.

There are three basic product operations in CGA, namely inner product, outer product and geometric product.
The inner product is used to verify the geometric properties like distances between two geometric primitives, denoted as $x$ $\cdot$ $y$, where $x$ and $y$ are multivectors \cite{bib28}.
In conformal space, the relationship between the distance of two vectors $p$, $q$ and inner product of two points $P$, $Q$ is shown as Eq. \eqref{equ2}.
\begin{equation}\label{equ2}
P \cdot Q =  - \frac{1}{2}{\left\| {p - q} \right\|^2}
\end{equation}
Outer product can construct geometry and extend dimension, denoted as $x$ $ \wedge$ $y$, where multivectors $x$ and $y$ are linearly independent.
Geometric product is represented by inner product and outer product, denoted as $xy = x \cdot y + x \wedge y$.
In CGA, the expressions of geometric primitives can be realized by inner product or outer product.
We describe some geometric expressions of geometric primitives in Table~\ref{tab2}.
\begin{table}[h]
\begin{center}
\begin{minipage}{370pt}
\caption{Geometric expressions based on CGA}\label{tab2}%
\begin{tabular}{|@{}|p{3.5cm}|l|l|@{}|}
\hline
Geometric primitives & Inner product expressions & Outer product expressions \\
\hline
Point $P$  \hspace{0.4cm} \vspace{0.2cm}  & $P = p + \frac{1}{2}{p^2}{e_\infty } + {e_0}$ &   \\
Sphere $S$  \hspace{0.4cm} \vspace{0.05cm}  & $S = P - \frac{1}{2}{r^2}{e_\infty }$ \hspace{0.4cm} \vspace{0.2cm}  & ${S^ * } = {P_1} \wedge {P_2} \wedge {P_3} \wedge {P_4}$  \\
Plane $\pi$  \hspace{0.4cm} \vspace{0.1cm}  & $\pi  = n + d{e_\infty }$  \hspace{0.4cm} \vspace{0.2cm} & ${\pi ^ * } = {P_1} \wedge {P_2} \wedge {P_3} \wedge {e_\infty }$  \\
Circle $Z$  \hspace{0.4cm} \vspace{0.1cm}  & $Z = {S_1} \wedge {S_2}$ \hspace{0.4cm} \vspace{0.2cm}  & ${Z^ * } = {P_1} \wedge {P_2} \wedge {P_3}$  \\
Line $L$  \hspace{0.4cm}  & $L = {\pi _1} \wedge {\pi _2}$  \hspace{0.4cm}  & ${L^ * } = {P_1} \wedge {P_2} \wedge {e_\infty }$  \\
\hline
\end{tabular}
\end{minipage}
\end{center}
\end{table}

According to Table~\ref{tab2}, in point $P$, $p$ represents the vector in three-dimensional Euclidean space.
In sphere $S$, $P$ represents the center of the sphere, and $r$ represents the radius.
In plane $\pi$, $n$ represents the unit normal vector, and $d$ represents the distance from the plane to the origin.
$Z$ is the circle obtained by the intersection of spheres $S_{1}$ and $S_{2}$.
$L$ is the line obtained by the intersection of planes $\pi_{1}$ and $\pi_{2}$.
The formal expressions of these five geometric primitives and the verifications of their related properties are available from the CGA library in HOL Light.
Here, the formal definition of the point $P$ in conformal space is shown in Definition \ref{def1}.

\begin{myDef}{Point}
\label{def1}
\end{myDef}
\begin{lstlisting}
&$\vdash$& !p. point_CGA p =
  p$1 % mbasis{1} + p$2 % mbasis{2} + p$3 % mbasis{3} +
  (&\&&1 / &\&&2 * (p dot p)) % null_inf + null_zero
\end{lstlisting}
In Definition \ref{def1}, the notation $\texttt {p\$1}$ represents the first component of the three-dimensional vector $p$.
The $\texttt {mbasis}$ function defines the basis blades indexed by subsets of (1..n).
For example, the basis vector $p_{1}$ can be represented as $\texttt {mbasis\{1\}}$.
$\texttt {null\_inf}$ represents the null vector $e_{\infty}$.
$\texttt {null\_zero}$ represents the null vector $e_{0}$.

Then, the formal definition of the sphere $S$ in conformal space is shown in Definition \ref{def2}.
\begin{myDef}{Sphere}
\label{def2}
\end{myDef}
\begin{lstlisting}
&$\vdash$& !p r. sphere_CGA p r =
  point_CGA p - ((&\&&1 / &\&&2) * (r pow 2)) % null_inf
\end{lstlisting}
In Definition \ref{def2}, $\tt p$ represents the center of the sphere $S$.
$\tt r$ represents the radius of the sphere $S$.
The formal expressions of the point $P$ and the sphere $S$ are used for the formalization of the closed balls and the capsules, which represent the components of the arms of the cooperative robot.

\section{Formalization of Geometric Models and Collision Conditions}\label{sec3}
In this section, we build a simplified robot model with geometric primitives.
We construct the shortest distance models and verify the collision conditions between geometric primitives by the formal method based on CGA.

\subsection{Formalization of Geometric Primitives Based on CGA}\label{subsec31}

\begin{figure}[h]
\centering
\includegraphics[width=0.7\columnwidth]{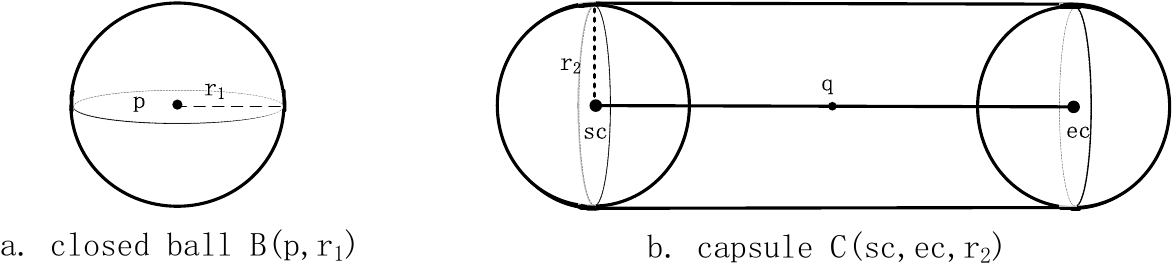}
\caption{Graphical representations of geometric primitives}\label{Figball}
\end{figure}

The robot body consists of joints and links, simplified into geometric primitives, such as closed balls and capsules.
Closed ball $B(p,r_{1})$ is shown in Fig.\ref{Figball}-a, and capsule $C(sc,ec,r_{2})$ is shown in Fig.\ref{Figball}-b.
Then, we construct formal models of closed balls and capsules and verify their related properties.

Mathematically, closed ball $B(p,r_{1})$ is defined as the set of specific points whose distance from the center $p$ of the closed ball $B(p,r_{1})$ is no larger than the radius $r_{1}$ of the closed ball $B(p,r_{1})$.
The corresponding formal representation of closed ball $B(p,r_{1})$ based on CGA is shown as Definition \ref{cball}.
\begin{myDef}{Closed ball}
\label{cball}
\end{myDef}
\begin{lstlisting}
&$\vdash$& !p r1.
  cball_CGA p r1 = {point_CGA x &$\mid$&
  &\&&0 <= (point_CGA x inner sphere_CGA p r1)$${} /\
  &\&&0 <= r1}
\end{lstlisting}
In Definition \ref{cball}, $\texttt {point\_CGA x}$ represents the point in closed ball $B(p,r_{1})$.
$\texttt {sphere\_CGA p r1}$  represents sphere $S$$(p,r_{1})$ with center $\tt p$ and radius $\tt r1$.
$\texttt {\$\$\{\}}$ represents the set index of multivectors, formally defined in HOL Light.
$\texttt {\&0 <= (point\_CGA x inner sphere\_}$ $\texttt {CGA p r1)\$\$\{\}}$ means that point $\tt x$ is not out of sphere $S$$(p,r_{1})$.
Notably, $\texttt {\&0 <= r1}$ represents the radius of the closed ball is not less than zero.
Therefore, Closed ball $B(p,r1)$ is the set of $\texttt {point\_CGA x}$.
According to Definition \ref{cball}, the distance between two points in the closed ball is bounded.
For example, the distance between point $\tt x$ and point $\tt y$ in closed ball $B(p,r_{1})$ is no larger than the diameter of closed ball $B(p,r_{1})$, shown as Theorem \ref{ballbou}.

\begin{myThm}{Boundness of the closed ball}
\label{ballbou}
\end{myThm}
\begin{lstlisting}
&$\vdash$& !x y p r1.
  point_CGA x IN cball_CGA p r1 /\
  point_CGA y IN cball_CGA p r1 ==> dist(x,y) <= &\&&2 * r1
\end{lstlisting}
The conditions $\texttt {point\_CGA x IN cball\_CGA p r1}$ and $\texttt {point\_CGA y IN cball\_CGA p r1}$ indicate that point $\tt x$ and point $\tt y$ are the points in closed ball $B(p, r1)$.
$\texttt {dist(x,y)}$ represents the distance between point $\tt x$ and point $\tt y$, and $\texttt {\&2 * r1}$ is the diameter of closed ball $B(p,r_{1})$.
We verify that the distance between any two points in the closed ball is bounded using the relevant theorems from the topology library in HOL Light.

Mathematically, center line $L(sc,ec)$ is defined as the set of points which are the centers of closed balls.
Capsule $C(sc,ec,r_{2})$ is defined as the set of closed balls whose centers move on center line $L(sc,ec)$.
$sc$ represents the starting point, $ec$ represents the ending point, $r_{2}$ represents the radius.
Then, we provide formal definitions of center line $L(sc,ec)$ and capsule $C(sc,ec,r_{2})$, shown as Definition \ref{centerline} and Definition \ref{capsule}.
\begin{myDef}{Center line}
\label{centerline}
\end{myDef}
\begin{lstlisting}
&$\vdash$& !sc ec.
  center_line_CGA sc ec =
  {point_CGA (sc + s % (ec - sc)) &$\mid$& &\&&0 <= s /\ s <= &\&&1}
\end{lstlisting}
In Definition \ref{centerline}, $\tt s$ represents the distance ratio that ranges from closed interval [0,1].
$\texttt {ec - sc}$ is the vector from $\tt sc$ to $\tt ec$.
$\texttt {point\_CGA (sc + s \% (ec - sc))}$ represent the points in $L(sc,ec)$.
Therefore, Center line $L(sc,ec)$ is the set of $\texttt {point\_CGA}$ $\texttt {(sc + s \% (ec-sc))}$.
\begin{myDef}{Capsule}
\label{capsule}
\end{myDef}
\begin{lstlisting}
&$\vdash$& !sc ec r2.
  capsule_CGA sc ec r2 =
  {point_CGA z &$\mid$& ?q. point_CGA q IN center_line_CGA sc ec
               /\ point_CGA z IN cball_CGA q r2}
\end{lstlisting}
In Definition \ref{capsule}, $\texttt {point\_CGA q}$ represents the point in center line $L(sc,ec)$.
$\texttt {point\_CGA}$ $\tt z$ represents the point in capsule $C(sc,ec,r_{2})$.
$\texttt {cball\_CGA q r2}$ represents the closed ball $B$$(q,r2)$ with center $\tt q$ and radius $\tt r2$.
Therefore, Capsule $C(sc,ec,r2)$ is the set of $\texttt {point\_CGA z}$ in $\texttt {cball\_CGA q r2}$, and the center $q$ is on $\texttt {center\_line\_CGA sc ec}$.
According to Definition \ref{centerline}, the distance between two points on center line $L(sc,ec)$ is bounded.
For example, the distance between point $z_{1}$ and point $z_{2}$ on center line $L(sc,ec)$ is no larger than the distance between the starting point $sc$ and the ending point $ec$, shown as Theorem \ref{cenbou}.
\begin{myThm}{Boundness of the center line}\label{cenbou}
\end{myThm}
\begin{lstlisting}
&$\vdash$& !z1 z2 sc ec.
  point_CGA z1 IN center_line_CGA sc ec /\ point_CGA z2 IN
  center_line_CGA sc ec ==> norm(point_CGA z1 inner point_CGA z2) <=
  norm(point_CGA sc inner point_CGA ec)
\end{lstlisting}
The condition $\texttt {point\_CGA z1 IN center\_line\_CGA sc ec}$ indicates that $\tt z1$ is the point on center line $C(sc,ec)$.
The condition $\texttt {point\_CGA z2 IN center\_}$ $\texttt {line\_CGA sc ec}$ indicates that $\tt z2$ is the point on center line $C(sc,ec)$.
$\texttt {norm(}$ $\texttt {point\_CGA z1}$ $\texttt {inner point\_CGA z2)}$ represents the distance between point $\tt z1$ and point $\tt z2$ on $L(sc,ec)$.
$\texttt {norm(point\_CGA sc inner}$ $\texttt {point\_CGA ec)}$ represents the distance between the starting point $\tt sc$ and the ending point $\tt ec$ of $L(sc,ec)$.
We simplify the subgoal using the distance theorem between two points from the CGA library in HOL Light.
The goal in Theorem \ref{cenbou} is verified by using the multiplication theorem of the norms and the related theorems of the absolute value of real numbers.

Then, we verify the relationship between the closed ball and the capsule based on their formal definitions.
When the starting point $sc$ and the ending point $ec$ overlap, center line $L(sc,ec)$ is equivalent to the point $sc$ or the point $ec$.
The corresponding representation is expressed in Theorem \ref{cencoin}.
\begin{myThm}{Equality of the center line}
\label{cencoin}
\end{myThm}
\begin{lstlisting}
&$\vdash$& !sc. center_line_CGA sc sc = {point_CGA sc}
&$\vdash$& !ec. center_line_CGA ec ec = {point_CGA ec}
\end{lstlisting}
$\texttt {\{point\_CGA sc\}}$ represents the starting point $\tt sc$.
$\texttt {\{point\_CGA ec\}}$ represents the ending point $\tt ec$.
We use the definition of the center line, the vector multiplication theorem, the vector addition theorem, and the vector subtraction theorem to simplify the subgoals.
Then, by using the theorem that the real number is not greater than itself, the two goals in Theorem \ref{cencoin} are proved.

When the starting point $sc$ and the ending point $ec$ overlap, the capsule $C(sc,ec,r_{2})$ is equivalent to the closed ball $B(sc,r_{2})$ or the closed ball $B(ec,r_{2})$.
The corresponding representation is expressed in Theorem \ref{capcoin}.
\begin{myThm}{Equality of the capsule}\label{capcoin}
\end{myThm}
\begin{lstlisting}
&$\vdash$& !sc r2. capsule_CGA sc sc r2 = cball_CGA sc r2
&$\vdash$& !ec r2. capsule_CGA ec ec r2 = cball_CGA ec r2
\end{lstlisting}
$\texttt {cball\_CGA sc r2}$ represents the closed ball $B(sc,r_{2})$.
$\texttt {cball\_CGA ec r2}$ represents the closed ball $B(ec,r_{2})$.
We simplify the subgoals using the formal definitions of the closed balls and the capsules.
Then, the goals in Theorem \ref{capcoin} are verified using the equality of the center line in Theorem \ref{cencoin} and the definition of the sphere in Definition \ref{def2}.

Through the analysis in Sect. \ref{subsec31}, we construct the formal models of geometric primitives and prove some basic properties of the closed ball, the center line, and the capsule by using tactics and theorem libraries in HOL Light.
These works lay the foundation for constructing the shortest distance models between geometric primitives in Sect. \ref{subsec32}.

\subsection{Formalization of the Shortest Distance Model}\label{subsec32}

\begin{figure}[h]
\centering
\includegraphics[width=0.9\columnwidth]{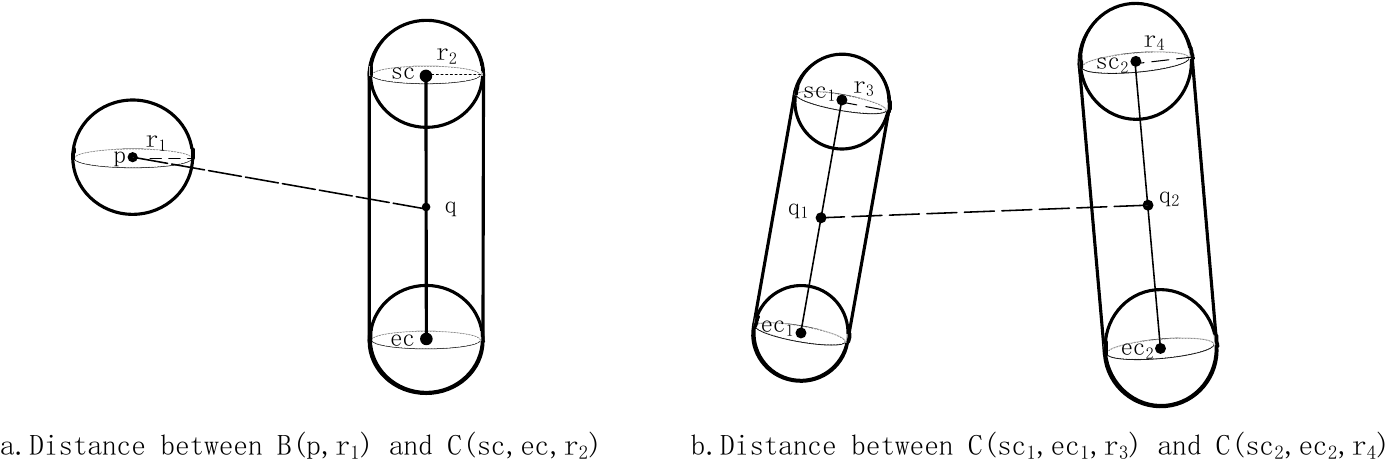}
\caption{Distance relationships between geometric primitives}\label{Figbacadis}
\end{figure}

Three types of collision between geometric primitives are closed ball-closed ball collision, closed ball-capsule collision and capsule-capsule collision.
Based on three types of collision, the shortest distance between geometric primitives is classified as the shortest distance between two closed balls, the shortest distance between the closed ball and the capsule, and the shortest distance between two capsules.
The distance relationships between geometric primitives are shown in Fig.\ref{Figbacadis}.

It is assumed that the shortest distance between two geometric primitives is $D$, and the shortest distance between the centers of closed balls (or center lines) is $d$.
$D$ is represented by $D - (R_{1} + R_{2})$.
$R_{1}$ and $R_{2}$ are the radii of two geometric primitives, respectively.
Since the radii of the two geometric primitives remain unchanged, the calculation of $D$ translates into the analysis of $d$.
Section \ref{subsec22} shows the relationship between inner product and distance in CGA.
We transform the distance calculation between two geometric primitives into calculating the inner product of points on geometric primitives.
Then, we construct the shortest distance models respectively.

Constructing the shortest distance model between two closed balls is equivalent to calculating the inner product between the centers of two closed balls.
The shortest distance model between two closed balls are formalized as Definition \ref{distball}.
\begin{myDef}{The shortest distance between two closed balls}
\label{distball}
\end{myDef}
\begin{lstlisting}
&$\vdash$& !p1 p2.
  center_dist_fa (point_CGA p1) (point_CGA p2) =
  &\&&2 * norm(point_CGA p1 inner point_CGA p2)
\end{lstlisting}
In Definition \ref{distball}, $\texttt {center\_dist\_fa}$ models the shortest distance between $\texttt {point\_}$ $\texttt {CGA p1}$ and $\texttt {point\_CGA p2}$.
$\texttt {point\_CGA p1}$ and $\texttt {point\_CGA p2}$ represent the centers of two closed balls respectively.
The result of the operation $\texttt {\&2 * norm}$ $\texttt {(point\_CGA p1 }$ $\texttt {inner point\_CGA p2)}$ is a real number.

Constructing the shortest distance model between closed ball $B(p,r_{1})$ and capsule $C(sc,ec,r_{2})$ is equivalent to calculating inner product between the center $p$ of closed ball $B(p,r_{1})$ and the point $q$ on center line $L(sc,ec)$.
The relationship of distance between closed ball $B(p,r_{1})$ and capsule $C(sc,ec,r_{2})$ is shown as Fig.\ref{Figbacadis}-a.
The shortest distance between closed ball $B(p,r_{1})$ and capsule $C(sc,ec,r_{2})$ is formalized as Definition \ref{distballcap}.
\begin{myDef}{The shortest distance between closed ball and capsule}
\label{distballcap}
\end{myDef}
\begin{lstlisting}
&$\vdash$& !p sc ec.
    center_dist_fb (point_CGA p) (center_line_CGA sc ec) =
    inf{&\&&2 * norm(point_CGA p inner point_CGA q) &$\mid$&
         point_CGA q IN center_line_CGA sc ec}
\end{lstlisting}
In Definition \ref{distballcap}, $\texttt {center\_dist\_fb}$ models the shortest distance between $\texttt {point\_}$ $\texttt {CGA p}$ and $\texttt {center\_line\_CGA sc ec}$.
$\tt inf$ is the infimum of the set of real numbers.
$\texttt {point\_CGA p}$ represents the center of the closed ball.
$\texttt {point\_CGA }$ $\texttt {q IN center\_line\_CGA sc ec}$ indicates that $\texttt {point\_CGA q}$ is the point on $\texttt {center\_line\_CGA sc ec}$.
$\texttt {\&2 * norm(point\_CGA p inner}$ $\texttt {point\_CGA q)}$ is a real number, and the shortest distance between the closed ball and capsule is the infimum of the set of real numbers.

According to Definition \ref{centerline}, we define the point $q$ on center line $L(sc,ec)$.
\begin{equation}\label{equ3}
q = sc + s * (ec - sc),0 \le s \le 1
\end{equation}
The distance between the center $p$ and point $q$ is expressed by a unary function $d(s)$, shown as Eq. \eqref{equ4}.
\begin{equation}\label{equ4}
d(s) = \sqrt {(ec - sc) \cdot (ec - sc){s^2} - 2s(p - sc) + (p - sc) \cdot (p - sc)}
\end{equation}
In Eq. \eqref{equ4}, $s$ represents the distance ratio $\frac{{\left\| {q - sc} \right\|}}{{\left\| {ec - sc} \right\|}}$ that ranges over the closed interval [0,1].
$ec - sc$ is the vector from $sc$ to $ec$, and $p - sc$ is the vector from $sc$ to $p$.

The shortest distance between closed ball $B(p,r_{1})$ and capsule $C(sc,ec,r_{2})$ is the minimum value of Eq. \ref{equ4}.
Then, we calculate the minimum value of Eq. \ref{equ4} by analyzing the relationship between $sc$ and $ec$, which is discussed in the following cases.

When $sc = ec$, the starting point $sc$ and the ending point $ec$ overlap.
According to Theorem \ref{capcoin},
the shortest distance between closed ball $B(p,r_{1})$ and capsule $C(sc,sc,r_{2})$ is equivalent to the shortest distance between closed ball $B(p,r_{1})$ and closed ball $B(sc,r_{2})$.

When $sc \ne ec$, the shortest distance between closed ball $B(p,r_{1})$ and capsule $C(sc,ec,r_{2})$ is discussed by situation depending on the position relationship between the closed ball and the capsule.
There are there cases which are listed below.
\begin{itemize}
  \item Case 1. $sc \ne ec$ $\wedge$ $(p - sc) \cdot (ec - sc) \le 0$ $\Rightarrow$
  $M_{1}$ = $\{$ d(s) $\mid$ s = 0$\}$
  \item Case 2. $sc \ne ec$ $\wedge$ $(ec - sc) \cdot (ec - sc) \le (p - sc) \cdot (ec - sc)$ $\Rightarrow$
  $M_{2}$ = $\{$ d(s) $\mid$ s = 1$\}$
  \item Case 3. $sc \ne ec$ $\wedge$ $0 < (p - sc) \cdot (ec - sc)$ $\wedge$ $(p - sc) \cdot (ec - sc) < (ec - sc) \cdot (ec - sc)$ $\Rightarrow$ $M_{3}$ = $\{$ d(s) $\mid$ s = $\frac{\|p - sc\|}{\|ec - sc\|}$$\}$
\end{itemize}
$M_{1}$, $M_{2}$ and $M_{3}$ represent the minimum values of Eq. \eqref{equ4} in each case.

According to three cases discussed above, we provide the formal representation for the minimum value of Eq. \eqref{equ4}, shown as Theorem \ref{mindiabc}.
\begin{myThm}{Minimum value under different conditions}
\label{mindiabc}
\end{myThm}
\begin{lstlisting}
&$\vdash$& !p r1 sc ec r2.
    center_dist_fb (point_CGA p) (center_line_CGA sc ec) =
    (if sc = ec
     then center_dist_fa (point_CGA p) (point_CGA sc)
     else if (p - sc) dot (ec - sc) <= &\&&0
     then &\&&2 * norm (point_CGA p inner point_CGA sc)
     else if (ec - sc) dot (ec - sc) <=
             (p - sc) dot (ec - sc)
     then &\&&2 * norm (point_CGA p inner point_CGA ec)
     else &\&&2 * norm (point_CGA p inner point_CGA (sc +
          (((p - sc) dot (ec - sc)) *
           inv ((ec - sc) dot (ec - sc))) % (ec - sc))))
\end{lstlisting}
When $(p - sc) \cdot (ec - sc) \le 0$, the angle between vector $p - sc$ and vector $ec - sc$ is a right angle or obtuse angle,
and the minimum value of Eq. \eqref{equ4} is equivalent to the distance between point $p$ and point $sc$.
When $(ec - sc) \cdot (ec - sc) \le (p - sc) \cdot (ec - sc)$, the angle between vector $p - sc$ and vector $ec - sc$ is an acute angle,
and the minimum value of Eq. \eqref{equ4} is equivalent to the distance between point $p$ and point $ec$.
When $0 < (p - sc) \cdot (ec - sc)$ and $(p - sc) \cdot (ec - sc) < (ec - sc) \cdot (ec - sc)$, the minimum value of Eq. \eqref{equ4} is equivalent to the distance between point $p$ and the point on center line $L(sc,ec)$ where $s$ is equal to $\frac{{\left\| {p - sc} \right\|}}{{\left\| {ec - sc} \right\|}}$.
The tactic $\mathsf {COND\_CASES\_TAC}$ from HOL Light is used to handle the $\texttt {if..else}$ statement in Theorem \ref{mindiabc}.
We simplify the subgoals using the shortest distance model between two closed balls and the formal definition of the center line.
Then, the goal in Theorem \ref{mindiabc} is verified by formal proof for three cases under the condition $sc \ne ec$.

Constructing the shortest distance model between capsule $C(sc_{1},ec_{1},r_{3})$ and capsule $C(sc_{2},ec_{2},r_{4})$ is equivalent to calculating inner product between the point $q_{1}$ on center line $L(sc_{1},ec_{1})$ and the point $q_{2}$ on center line $L(sc_{2},ec_{2})$.
The relationship of the distance between capsule $C(sc_{1},ec_{1},r_{3})$ and capsule $C(sc_{2},ec_{2},r_{4})$ is shown as Fig.\ref{Figbacadis}-b.
The shortest distance between capsule $C(sc_{1},ec_{1},r_{3})$ and capsule $C(sc_{2},ec_{2},r_{4})$ is formalized as Definition \ref{distcapcap}.
\begin{myDef}{The shortest distance between two capsules}
\label{distcapcap}
\end{myDef}
\begin{lstlisting}
&$\vdash$& !sc1 ec1 sc2 ec2.
    center_dist_fc (center_line_CGA sc1 ec1)
                   (center_line_CGA sc2 ec2) =
    inf {&\&&2 * norm(point_CGA q1 inner point_CGA q2) &$\mid$&
         point_CGA q1 IN center_line_CGA sc1 ec1 /\
         point_CGA q2 IN center_line_CGA sc2 ec2}
\end{lstlisting}
In Definition \ref{distcapcap}, $\texttt {center\_dist\_fc}$ models the shortest distance between $\texttt {center\_line\_CGA sc1}$ $\texttt { ec1}$ and $\texttt {center\_line\_CGA sc2 ec2}$.
$\texttt {center\_line\_CGA sc1 ec1}$ represents the center line $L(sc_{1},ec_{1})$ of capsule $C(sc_{1},ec_{1},r_{3})$ with starting point $\tt sc1$ and ending point $\tt ec1$.
$\texttt {center\_line\_CGA sc2 ec2}$ represents the center line $L(sc_{2},ec_{2})$ of capsule $C(sc_{2},ec_{2},r_{4})$ with starting point $\tt sc2$ and ending point $\tt ec2$.
$\texttt {point\_CGA q1}$ represents the point on $L(sc_{1},ec_{1})$.
$\texttt {point\_CGA q2}$ represents the point on $L(sc_{2},ec_{2})$.
The shortest distance between two capsules is the infimum of the set of $\texttt {\&2 * norm(point\_CGA q1 inner point\_CGA q2)}$.

According to Definition \ref{centerline}, we define the point $q_{1}$ on the center line $L(sc_{1},ec_{1})$ and the point $q_{2}$ on the center line $L(sc_{2},ec_{2})$ respectively, shown as Eq. \eqref{equ5} and Eq. \eqref{equ6}.
\begin{equation}\label{equ5}
q_{1} = sc_{1} + s_{1} * (ec_{1} - sc_{1}),0 \le s_{1} \le 1
\end{equation}
\begin{equation}\label{equ6}
q_{2} = sc_{2} + s_{2} * (ec_{2} - sc_{2}),0 \le s_{2} \le 1
\end{equation}
$s_{1}$ represents the distance ratio $\frac{{\left\| {q_{1} - sc_{1}} \right\|}}{{\left\| {ec_{1} - sc_{1}} \right\|}}$ that ranges from closed interval [0,1].
$ec_{1 }- sc_{1}$ represents the vector from the starting point $sc_{1}$ to the ending point $ec_{1}$.
$s_{2}$ represents the distance ratio $\frac{{\left\| {q_{2} - sc_{2}} \right\|}}{{\left\| {ec_{2} - sc_{2}} \right\|}}$ that ranges from closed interval [0,1].
$ec_{2} - sc_{2}$ represents the vector from the starting point $sc_{2}$ to the ending point $ec_{2}$.

The distance between point $q_{1}$ and point $q_{2}$ is expressed by a binary function $d(s_{1},s_{2})$, shown as Eq. \eqref{equ7}.
\begin{equation}\label{equ7}
d(s_{1},s_{2}) = \sqrt {as{_{1}^2} + cs{_{2}^2} - 2bs_{1}s_{2} + 2ds_{1} - 2es_{2} + f}
\end{equation}
In Eq. \eqref{equ7}, $u = ec_{1} - sc_{1}, v = ec_{2} - sc_{2}, w = sc_{1} - sc_{2}$, $a = u \cdot u, b = - (u \cdot v), c = v \cdot v, d = u \cdot w, e = - (v \cdot w), f = (w \cdot w)$.
$a$, $b$, $c$, $d$, $e$ and $f$ are real numbers.
$u$, $v$ and $w$ are vectors.
The shortest distance between $C(sc_{1},ec_{1},r_{3})$ and $C(sc_{2},ec_{2},r_{4})$ is equivalent to the minimum value of Eq. \eqref{equ7}.
We calculate the binary function described in Eq. \eqref{equ7} by dividing it into two situations.
One situation is that the stationary point exists.
The other situation is that the stationary point does not exist.
Then, we verify the minimum value in the two situations above.

When the stationary point exists,
it is expressed by $s_{1}{_k} = \frac{{be - cd}}{{ac - {b^2}}}$ and $s_{2}{_k} = \frac{{bd - ae}}{{ac - {b^2}}}$.
$s_{1}{_k}$ and $s_{2}{_k}$ are meaningful where $ac - {b^2} \ne 0$.
According to Eq. \eqref{equ5} and Eq. \eqref{equ6}, $s{1_m}$ and $s{2_m}$ range from closed interval [0,1].
Under the condition of $ac - {b^2} \ne 0$ and $s_{1}{_k}$,$s_{2}{_k} \in [0,1]$, the minimum value of Eq. \eqref{equ7} is the calculation result of binary function where $s_{1} = s_{1}{_k}$ and $s_{2} = s_{2}{_k}$.
This property is shown in Theorem \ref{theopse3}.
\begin{myThm}{Minimum value if $ac - {b^2} \ne 0$ and $s_{1}{_k} , s_{2}{_k} \in [0,1]$}
\label{theopse3}
\end{myThm}
\begin{lstlisting}
&$\vdash$& !sc1 ec1 sc2 ec2 u v w a b c d e f s1k s2k.
    u = ec1 - sc1 /\ v = ec2 - sc2 /\ w = sc1 - sc2 /\
    a = u dot u /\ b = --(v dot u) /\ c = v dot v /\
    d = u dot w /\ e = --(v dot w) /\ f = w dot w /\
    s1k = (b * e - c * d) * inv (a * c - b pow 2) /\
    s2k = (b * d - a * e) * inv (a * c - b pow 2) /\
    ~(a * c - b pow 2 = &\&&0) /\
    &\&&0 <= s1k /\ s1k <= &\&&1 /\ &\&&0 <= s2k /\ s2k <= &\&&1
    ==> center_dist_fc (center_line_CGA sc1 ec1)
                       (center_line_CGA sc2 ec2) =
        &\&&2 * norm(point_CGA (sc1 + s1k % (ec1 - sc1)) inner
                  point_CGA (sc2 + s2k % (ec2 - sc2)))
\end{lstlisting}
$\texttt {b pow 2}$ is the square of $\tt b$.
$\texttt {point\_CGA (sc1 + s1k \% (ec1 - sc1))}$ represents the point in center line $L(sc_{1},ec_{1})$ where $s_{1} = s_{1}{_k}$.
$\texttt {point\_CGA (sc2 +}$ $\texttt {s2k \% (ec2 - sc2))}$ represents the point in center line $L(sc_{2},ec_{2})$ where $s_{2} = s_{2}{_k}$.
There are many preconditions in Theorem \ref{theopse3}.
We use the tactic $\mathsf {CLAIM\_TAC}$ in HOL Light to decompose the goal into multiple subgoals and solve them separately.

When the stationary point does not exist, there are nine cases.
Then, we analyze the nine cases separately and calculate the minimum value of Eq. \eqref{equ7}.
The nine cases are listed below.
\begin{itemize}
  \item Case 1. $ec_{1} = sc_{1}$ $\wedge$ $sc_{2} \ne ec_{2}$ $\Rightarrow$ $N_{1}$ = $inf$ $\{$ $d(s_{1},s_{2})$ $\mid$ $s_{1} = 0$, $0 \le s_{2} \le 1$ $\}$
  \item Case 2. $sc_{1} = ec_{1}$ $\wedge$ $sc_{2} \ne ec_{2}$ $\Rightarrow$ $N_{2}$ = $inf$ $\{$ $d(s_{1},s_{2})$ $\mid$ $s_{1} = 1$, $0 \le s_{2} \le 1$ $\}$
  \item Case 3. $ec_{2} = sc_{2}$ $\wedge$ $sc_{1} \ne ec_{1}$ $\Rightarrow$ $N_{3}$ = $inf$ $\{$ $d(s_{1},s_{2})$ $\mid$ $s_{2} = 0$, $0 \le s_{1} \le 1$ $\}$
  \item Case 4. $sc_{2} = ec_{2}$ $\wedge$ $sc_{1} \ne ec_{1}$ $\Rightarrow$ $N_{4}$ = $inf$ $\{$ $d(s_{1},s_{2})$ $\mid$ $s_{2} = 1$ , $0 \le s_{1} \le 1$ $\}$
  \item Case 5. $N_{1} \wedge N_{2} \wedge N_{3} \wedge N_{4} \wedge s_{1k} < 0 \Rightarrow inf$ $(N_{1} \cup N_{2} \cup N_{3} \cup N_{4})$
  \item Case 6. $N_{1} \wedge N_{2} \wedge N_{3} \wedge N_{4} \wedge s_{1k} > 1 \Rightarrow inf$ $(N_{1} \cup N_{2} \cup N_{3} \cup N_{4})$
  \item Case 7. $N_{1} \wedge N_{2} \wedge N_{3} \wedge N_{4} \wedge s_{2k} < 0 \Rightarrow inf$ $(N_{1} \cup N_{2} \cup N_{3} \cup N_{4})$
  \item Case 8. $N_{1} \wedge N_{2} \wedge N_{3} \wedge N_{4} \wedge s_{2k} > 1 \Rightarrow inf$ $(N_{1} \cup N_{2} \cup N_{3} \cup N_{4})$
  \item Case 9. $N_{1} \wedge N_{2} \wedge N_{3} \wedge N_{4} \wedge ac - {b^2} = 0 \Rightarrow inf$ $(N1 \cup N_{2} \cup N_{3} \cup N_{4})$
\end{itemize}
$N_{1}$ is the minimum value of Eq. \eqref{equ7} in Case 1.
$N_{2}$ is the minimum value of Eq. \eqref{equ7} in Case 2.
$N_{3}$ is the minimum value of Eq. \eqref{equ7} in Case 3.
$N_{4}$ is the minimum value of Eq. \eqref{equ7} in Case 4.
From Case 1 to Case 4, $s_{1}$ and $s_{2}$ take values on the boundaries of the closed interval [0,1].
From Case 5 to Case 8, $s_{1}{_k}$ and $s_{2}{_k}$ are not all in the closed interval [0,1], the stationary point violates Eq. \eqref{equ5} and Eq. \eqref{equ6}.
In Case 9, $ac - {b^2} = 0$ represents that center line $L(sc_{1},ec_{1})$ is parallel to center line $L(sc_{2},ec_{2})$.
The process of verification is similar in the nine cases.
We take Case 1 as an example, shown as Theorem \ref{disL1}.
\begin{myThm}{Minimum value in Case 1}
\label{disL1}
\end{myThm}
\begin{lstlisting}
&$\vdash$& !sc1 ec1 sc2 ec2 r3 r4.
    ~(sc2 = ec2)
    ==> center_dist_fc (center_line_CGA sc1 sc1)
                       (center_line_CGA sc2 ec2) =
        (if (ec2 - sc2) dot (sc1 - sc2) <= &\&&0
         then &\&&2 * norm (point_CGA sc1 inner point_CGA sc2)
         else if (ec2 - sc2) dot (ec2 - sc2) <=
                 (ec2 - sc2) dot (sc1 - sc2)
         then &\&&2 * norm (point_CGA sc1 inner point_CGA ec2)
         else &\&&2 * norm (point_CGA sc1 inner
              point_CGA (sc2 + (((ec2 - sc2) dot
                        (sc1 - sc2)) * inv ((ec2 - sc2) dot
                        (ec2 - sc2))) % (ec2 - sc2))))
\end{lstlisting}
$N_{1}$ is equal to the infimum of the set $\{$ $d(s_{1},s_{2})$ $\mid$ $s_{1} = 0$, $0 \le s_{2} \le 1$ $\}$.
In Theorem \ref{disL1}, under the precondition $\texttt {$\sim$(sc2 = ec2)}$, the minimum value of Equation \eqref{equ5} is expressed by $\texttt {center\_dist\_fc (center\_line\_CGA sc1 sc1) (center\_line\_CGA sc2 ec2)}$ where $ec_{1} = sc_{1}$.
We divide $0 \le s_{2} \le 1$ into three conditions which include $s_{2} = 0$, $s_{2} = 1$, and $s_{2} = \frac{{\left\| {sc_{1} - sc_{2}} \right\|}}{{\left\| {ec_{2} - sc_{2}} \right\|}}$.
For example, the condition $\texttt {(ec2 - sc2) dot (sc1 - sc2) <= \&0}$ is satisfied when $s_{2} = 0$, and the minimum value is equal to the distance between $\texttt {point\_CGA sc1}$ and $\texttt {point\_CGA sc2}$.
We prove the three cases under the conditions $ec_{1} = sc_{1}$ and $sc_{2}$ $\ne$ $ec_{2}$.
In verifying the three cases separately, we need to use the closed ball-closed ball shortest distance model, the closed ball-capsule shortest distance model, and the formal definition of the center line.
The theorems from the vector library and the real library in HOL Light are used to verify the three cases.
Then, we use the proof results for these three cases to verify Theorem \ref{disL1} in combination with the tactic $\mathsf {REAL\_ARITH}$ in HOL Light.

\subsection{Formal Verification of Geometric Collision Conditions}
\label{subsec33}
Three types of collision between geometric primitives are closed ball-closed ball collision, closed ball-capsule collision and capsule-capsule collision.
Based on these three types of collision, the shortest distance between geometric primitives can be classified as the shortest distance between two closed balls, the shortest distance between the closed ball and the capsule, and the shortest distance between two capsules.
Then, we provide formal verification of the collision conditions of two closed balls, the closed ball and the capsule, and two capsules.

According to Sect. \ref{subsec32},
the relationship between $D$ and $d$ is $D = d -(R_{1} + R_{2})$.
If $D \leq 0$, which means $d \leq R_{1} + R_{2}$, it can be judged as collision of geometric primitives.
If $D > 0$, which means $d > R_{1} + R_{2}$, it can be judged as the separation of geometric primitives.
Then, we verify that geometric primitives collide by analyzing that $d$ is not larger than the sum of the radii of geometric primitives.
According to Definition \ref{distball}, we verify the collision conditions between closed ball $B(p_{1},r_{3})$ and closed ball $B(p_{2},r_{4})$, shown as Theorem \ref{babaco}.
\begin{myThm}{Collision of two closed balls}
\label{babaco}
\end{myThm}
\begin{lstlisting}
&$\vdash$& !p1 p2 r3 r4.
    &\&&0 <= r3 /\ &\&&0 <= r4
    ==> (~DISJOINT (cball_CGA p1 r3) (cball_CGA p2 r4) <=>
        center_dist_fa (point_CGA p1) (point_CGA p2)
        <= (r3 + r4) pow 2)
\end{lstlisting}
The preconditions $\texttt {\&0 <= r3}$ and $\texttt {\&0 <= r4}$ indicate that the closed ball $B(p_{1},r_{3})$ and the closed ball $B(p_{2},r_{4})$ are not empty set.
When the shortest distance between closed ball $B(p_{1},r_{3})$ and closed ball $B(p_{2},r_{4})$ is no larger than the sum of the radii, two closed balls collide.

According to Definition \ref{distballcap}, we verify the collision conditions between closed ball $B(p,r_{1})$and capsule $C(sc,ec,r_{2})$.
The collision conditions between a closed ball and capsule are similar to those between two closed balls.
The collision model of the closed ball and the capsule is shown as Theorem \ref{bacaco}.
\begin{myThm}{Collision of closed ball and capsule}
\label{bacaco}
\end{myThm}
\begin{lstlisting}
&$\vdash$& !p r1 sc ec r2.
    &\&&0 <= r1 /\ &\&&0 <= r2 ==>
    (~DISJOINT (cball_CGA p r1) (capsule_CGA sc ec r2) <=>
    center_dist_fb (point_CGA p) (center_line_CGA sc ec) <=
    (r1 + r2) pow 2)
\end{lstlisting}
The preconditions $\texttt {\&0 <= r1}$ and $\texttt {\&0 <= r2}$ indicate that the closed ball $B(p,r_{1})$ and the capsule $C(sc,ec,r_{2})$ are not empty set.
When the shortest distance between closed ball $B(p,r_{1})$ and capsule $C(sc,ec,r_{2})$ is no larger than the sum of the radii, the closed ball and capsule collide.

According to Definition \ref{distcapcap}, we verify the collision conditions between capsule $C(sc_{1},ec_{1},r_{3})$ and capsule $C(sc_{2},ec_{2},r_{4})$.
Compared with collision conditions between closed ball and capsule, more preconditions are needed for two capsules.
Then, we verify the collision conditions between two capsules, shown as Theorem \ref{cacaco}.
\begin{myThm}{Collision of two capsules}
\label{cacaco}
\end{myThm}
\begin{lstlisting}
&$\vdash$& !r3 r4 sc1 ec1 sc2 ec2 u v w a b c d e f s1k s2k.
    &\&&0 <= r3 /\ &\&&0 <= r4 /\
    u = ec1 - sc1 /\ v = ec2 - sc2 /\ w = sc1 - sc2 /\
    a = u dot u /\ b = --(v dot u) /\ c = v dot v /\
    d = u dot w /\ e = --(v dot w) /\ f = w dot w /\
    s1k = (b * e - c * d) * inv (a * c - b pow 2) /\
    s2k = (b * d - a * e) * inv (a * c - b pow 2) /\
    ~(a * c - b pow 2 = &\&&0) /\
    &\&&0 <= s1k /\ s1k<= &\&&1 /\ &\&&0 <= s2k /\ s2k <= &\&&1 ==>
    (~DISJOINT (capsule_CGA sc1 ec1 r3)
               (capsule_CGA sc2 ec2 r4)
    <=> center_dist_fc (center_line_CGA sc1 ec1)
        (center_line_CGA sc2 ec2) <= (r3 + r4) pow 2)
\end{lstlisting}
The preconditions $\texttt {\&0 <= r3}$ and $\texttt {\&0 <= r4}$ indicate that the capsule $C(sc_{1},ec_{1},r_{3})$ and the capsule $C(sc_{2},ec_{2},r_{4})$ are not empty set.
The precondition that the stationary point in Eq. \eqref{equ5} exists must be satisfied.
And the center line of capsule $C(sc_{1},ec_{1},r_{3})$ is not parallel to the center line of capsule $C(sc_{2},ec_{2},r_{4})$.
Under these preconditions, we verify that two capsules collide when the shortest distance between the capsule $C(sc_{1},ec_{1},r_{3})$ and the capsule $C(sc_{2},ec_{2},r_{4})$ is no larger than the sum of the radii.

Based on the analysis and verification above, we construct a formal verification framework for the collision detection method based on the models of geometric primitives.

\section{Higher-Order Expressions of Robot Collision Detection}\label{sec4}
In this section, we use the formal verification framework of the collision detection method constructed in Sect. \ref{sec3} to implement collision detection between robots.
The process of robot collision detection can be divided into two major parts.
First, we implement a higher-order logic representation of the robot body based on the models of geometric primitives.
Second, we verify the robot collision model based on the shortest distance models and collision detection conditions.
Then, we apply the formal verification framework to collision detection between the two single-arm industrial cooperative robots.
\subsection{Formal Modeling of Robot Collision Model}\label{subsec41}
The basic components of the robot are separable.
Therefore, the geometric model of the robot body can be represented by the union of several geometric primitives.
Based on iterate operation and UNION operation of the set library in HOL Light, the geometric model of the robot body is formalized as Definition \ref{robot}.
\begin{myDef}{The robot body}
\label{robot}
\end{myDef}
\begin{lstlisting}
&$\vdash$& robot = iterate (UNION)
\end{lstlisting}
In Definition \ref{robot}, $\tt iterate$ is the iteration of the binary operation, which is expressed as $\texttt {iterate op s f}$.
$\tt op$ is the binary operation, and the binary operation used in Definition \ref{robot} is $\tt UNION$.
The representation in the complete meaning of $\tt robot$ is $\texttt {robot s f}$.
$\tt s$ represents the set of indices of the components, which can be represented by the natural number segment $\texttt {(m..n)}$.
$\tt m$ and $\tt n$ denote the indices of the robot components.
$\tt f$ represents the mapping relationship.
Therefore, we can model a robot that maps $\tt f$ to the indices set $\texttt {(m..n)}$ which represent the indices of the components (geometric primitives), and performs $\tt UNION$ operation on the geometric primitives (the closed balls and the capsules).
Then, depending on the different values of m and n, we verify the configuration rules of the robot.
The verification of the configuration rules of the robot is shown as Theorem \ref{robotexi}.
\begin{myThm}{The configuration rules of the robots}
\label{robotexi}
\end{myThm}
\begin{lstlisting}
&$\vdash$& (!m f. robot (m..0) f =
    (if m = 0 then f(0) else {})) /\
    (!m n f. robot (m..SUC n) f =
    (if m <= SUC n then robot (m..n) f UNION f (SUC n)
     else robot (m..n) f))
\end{lstlisting}
$\texttt {SUC n}$ denotes a natural number $n+1$.
The configuration rules of the robots can be divided into two types.
If $\tt m$ is equal to zero, $\texttt {robot (m..0) f}$ represents the robot which has only one component $\texttt {f(0)}$, or $\texttt {robot (m..0) f}$ is the empty set.
If $\texttt {SUC n}$ does not belong to the set $\texttt {(m..n)}$ of indices of the components, the original set $\texttt {(m..n)}$ will be expanded to a UNION set $\texttt {(m..SUC n)}$ with $\texttt {SUC n}$, or $\texttt {robot (m..0) f}$ is expressed by $\texttt {robot (m..n) f}$.
In addition, if $\tt m$ is equal to $\tt n$, $\tt s$ is expressed as $\texttt {(n..n)}$, and $\texttt {robot (n..n) f}$ denotes a component of robot with index $\tt n$.
We verify this property in Theorem \ref{sincoro}.
\begin{myThm}{The single component of the robot}
\label{sincoro}
\end{myThm}
\begin{lstlisting}
&$\vdash$& !f n. robot (n..n) f = f(n)
\end{lstlisting}
Depending on the value of $\tt n$, $\texttt {f(n)}$ represents different components of the robot body.
According to Definition \ref{robot} and Theorem \ref{robotexi}, if the function values of two different functions (mapping relation) $\tt f$ and $\tt g$ are the same under the same component indices set $\tt s$, the two robots $\texttt {robot s f}$ and $\texttt {robot s g}$ are also the same robot.
Therefore, we verify the equivalence condition for two robots in Theorem \ref{equco}.
\\
\begin{myThm}{Equivalent conditions of two robots}
\label{equco}
\end{myThm}
\begin{lstlisting}
&$\vdash$& !f g s.
  (!x. x IN s ==>
  (f(x) = g(x))) ==> (robot s f = robot s g)
\end{lstlisting}
The mapping relations $\tt f$ and $\tt g$ contain not only the indices information of the components, but also the type and pose information of the components of the corresponding indices.
Therefore, the formal representation of the robot body contains the information of the types, numbers, indices of the robot components.

The geometric model of the robot body is composed of geometric primitives.
Notably, there are intersections between the components of the robot.
Based on the geometric model of the robot body, we give a formal description of the robot collision model in Definition \ref{rocomo}.
\begin{myDef}{The robot collision model}
\label{rocomo}
\end{myDef}
\begin{lstlisting}
&$\vdash$& !n f m g.
  collision (n,f) (m,g) <=>
  ~(robot (0..n) f INTER robot (0..m) g = {})
\end{lstlisting}
In Definition \ref{rocomo}, $\texttt {robot (0..n) f INTER robot (0..m) g = \{\}}$ represents the intersection of $\texttt {robot (0..n) f}$ and $\texttt {robot (0..m) g}$ is empty.
Therefore, we can conclude that two robots collide when the intersection of their components is not empty.
Then, we verify the equivalent collision conditions of two robots, shown as Theorem \ref{robotcoco}.
\begin{myThm}{Equivalent collision conditions of two robots}
\label{robotcoco}
\end{myThm}
\begin{lstlisting}
&$\vdash$& !n m f g.
    collision (n,f) (m,g) <=>
    ~(!i j. 1 <= i /\ i <= SUC n /\ 1 <= j /\ j <= SUC m
    ==> DISJOINT (f (i - 1)) (g (j - 1)))
\end{lstlisting}
$\texttt {1 <= i}$, as well as $\texttt {1 <= j}$, represents that the robot has one component at least.
$\texttt {i <= SUC n}$, as well as $\texttt {j <= SUC n}$, represents that the robot has $\texttt {n+1}$ components at most.
In the verification process of Theorem \ref{robotcoco}, tactic $\mathsf  {ONCE\_REWRITE\_TAC}$ in HOL Light is needed.
After the above analysis, we construct the robot collision model and verify the equivalent collision conditions of the robots.

\subsection{Verification of Collision Detection for two Single-Arm Industrial Cooperative Robots} \label{subsec42}

\begin{figure}[h]
\centering
\includegraphics[width=0.65\columnwidth]{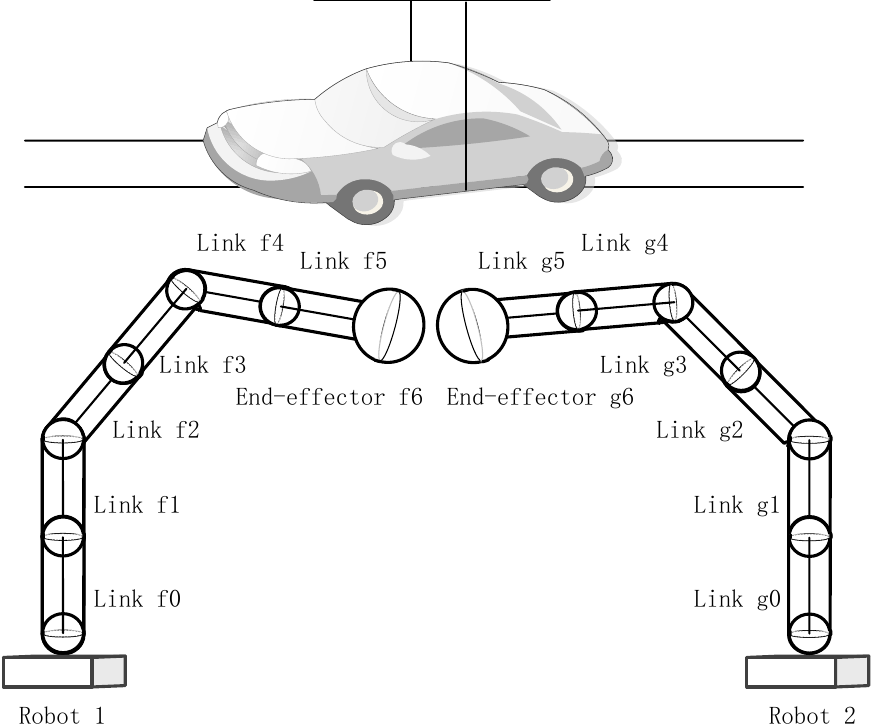}
\caption{Two single-arm industrial cooperative robots}\label{Figassrobot}
\end{figure}

To verify that the proposed formal verification framework of the robot collision detection method is available, we apply the framework to the collision detection for two single-arm industrial cooperative robots.
Two single-arm cooperative robots are shown in Fig.\ref{Figassrobot}.
The arm of the cooperative robot consists of six links and an end effector.
The links can be abstracted as capsules.
And the end effector can be abstracted as a closed ball.
We numbered the links and end effectors for two single-arm industrial cooperative robots.

According to the formal definition of the robot in Definition \ref{robot} and the configuration rules of the robot in Theorem \ref{robotexi}, we can formalize the $Robot$ $1$ as $\texttt {robot (0..6) f}$, and formalize the $Robot$ 2 as $\texttt {robot (0..6) g}$.
Then, by using the formal models of the two robots, we verify whether the two robots in Fig.\ref{Figassrobot} collide or not.
The formal verification of this property is shown as Theorem \ref{tworobotnoco}.
\\
\\
\begin{myThm}{Collision of two single-arm industrial cooperative robots}
\label{tworobotnoco}
\end{myThm}	
\begin{lstlisting}
&$\vdash$& !f g.
  f(0) = capsule_CGA a1 b1 r2 /\
  f(1) = capsule_CGA b1 c1 r2 /\
  f(2) = capsule_CGA c1 d1 r2 /\
  f(3) = capsule_CGA d1 e1 r2 /\
  f(4) = capsule_CGA e1 x1 r2 /\
  f(5) = capsule_CGA x1 y1 r2 /\
  f(6) = cball_CGA y1 r1 /\
  g(0) = capsule_CGA a2 b2 r2 /\
  g(1) = capsule_CGA b2 c2 r2 /\
  g(2) = capsule_CGA c2 d2 r2 /\
  g(3) = capsule_CGA d2 e2 r2 /\
  g(4) = capsule_CGA e2 x2 r2 /\
  g(5) = capsule_CGA x2 y2 r2 /\
  g(6) = cball_CGA y2 r1 /\
  &\&&0 <= r1 /\ &\&&0 <= r2
  ==> ((collision (6,f) (6,g)) <=>
      (~(!i j. 1 <= i /\ i <= SUC 6 /\ 1 <= j /\ j <= SUC 6
      ==> (DISJOINT (f(i - 1)) (g(j - 1)))) \/
          (center_dist_fb (point_CGA y2)
          (center_line_CGA a1 b1) <= (r1 + r2) pow 2) \/
          (center_dist_fb (point_CGA y2)
          (center_line_CGA b1 c1) <= (r1 + r2) pow 2) \/
          (center_dist_fb (point_CGA y2)
          (center_line_CGA c1 d1) <= (r1 + r2) pow 2) \/
          (center_dist_fb (point_CGA y2)
          (center_line_CGA d1 e1) <= (r1 + r2) pow 2) \/
          (center_dist_fb (point_CGA y2)
          (center_line_CGA e1 x1) <= (r1 + r2) pow 2) \/
          (center_dist_fb (point_CGA y2)
          (center_line_CGA x1 y1) <= (r1 + r2) pow 2) \/
          (center_dist_fb (point_CGA y1)
          (center_line_CGA a2 b2) <= (r1 + r2) pow 2) \/
          (center_dist_fb (point_CGA y1)
          (center_line_CGA b2 c2) <= (r1 + r2) pow 2) \/
          (center_dist_fb (point_CGA y1)
          (center_line_CGA c2 d2) <= (r1 + r2) pow 2) \/
          (center_dist_fb (point_CGA y1)
          (center_line_CGA d2 e2) <= (r1 + r2) pow 2) \/
          (center_dist_fb (point_CGA y1)
          (center_line_CGA e2 x2) <= (r1 + r2) pow 2) \/
          (center_dist_fb (point_CGA y1)
          (center_line_CGA x2 y2) <= (r1 + r2) pow 2) \/
          (center_dist_fa (point_CGA y1) (point_CGA y2) <=
          (r1 + r1) pow 2)))
\end{lstlisting}
In Theorem \ref{tworobotnoco}, $\texttt {f(0)}$, $\texttt {f(1)}$, $\texttt {f(2)}$, $\texttt {f(3)}$, $\texttt {f(4)}$, and $\texttt {f(5)}$ represent the six links of the $Robot$ 1 respectively.
$\texttt {f(6)}$ represents the end effector of the $Robot$ 1.
$\texttt {g(0)}$, $\texttt {g(1)}$, $\texttt {g(2)}$, $\texttt {g(3)}$, $\texttt {g(4)}$, and $\texttt {g(5)}$ represent the six links of the $Robot$ 2 respectively.
$\texttt {g(6)}$ represents the end effector of the $Robot$ 2.
$\texttt {\&0 <= r1}$ and $\texttt {\&0 <= r2}$ represent closed balls and capsules which represent the end effectors and the links of two single-arm industrial cooperative robots are not empty sets.
According to Definition \ref{rocomo}, $\texttt {collision (6,f) (6,g)}$ is equivalent to $\texttt {$\sim$(robot (0..6) f INTER robot (0..6) g = \{\})}$.

If there is an intersection between any component (the closed ball or the capsule) in $Robot$ 1 and any component (the closed ball or the capsule) in $Robot$ 2, we can conclude that these two robots collide.
Two single-arm industrial cooperative robots are formally defined as $\texttt {robot (0..6) f}$ and $\texttt {robot (0..6) g}$.
Therefore, the equivalent collision conditions, shown in Theorem \ref{robotcoco}, can be used to verify whether two single-arm industrial cooperative robots collide.
Based on the equivalent collision conditions, Theorem \ref{tworobotnoco} is verified using the shortest distance models and the collision conditions of geometric primitives.
Therefore, we provide a general model for verifying the collision between two single-arm industrial cooperative robots using the constructed formal verification framework in Sect. \ref{sec3}.

Then, based on the general model which is provided in Theorem \ref{tworobotnoco}, we verify that two single-arm industrial cooperative robots, which are given the specific position parameters, collide.
According to Fig.\ref{Figassrobot} and the formal definitions of two single-arm industrial cooperative robots, we provide the parameter list for $Robot$ 1 and $Robot$ 2, shown as Table \ref{tab3}.
\begin{table}[h]
\begin{center}
\begin{minipage}{440pt}
\caption{Parameter list of $Robot$ 1 and $Robot$ 2}\label{tab3}%
\begin{tabular}{|@{}|p{3cm}|p{3.5cm}|l|l|l|@{}|}
\hline
Components & Geometric primitives & Starting points & Ending points & {Radius} \\
\hline
Link f(0) & Capsule  & $a_{1}(0,0,0)$ & $b_{1}(0,0,65)$ & $r_{2} = 15mm$ \\
Link f(1) & Capsule  & $b_{1}(0,0,65)$ & $c_{1}(0,0,130)$ & $r_{2} = 15mm$ \\
Link f(2) & Capsule  & $c_{1} (0,0,130)$ & $d_{1} (35,0,130)$ & $r_{2} = 15mm$ \\
Link f(3) & Capsule  & $d_{1} (35,0,130)$ & $e_{1} (70,0,130)$ & $r_{2} = 15mm$ \\
Link f(4) & Capsule  & $e_{1} (70,0,130)$ & $x_{1} (70,45,130)$ & $r_{2} = 15mm$ \\
Link f(5) & Capsule  & $x_{1} (70,45,130)$ & $y_{1} (70,90,130)$ & $r_{2} = 15mm$ \\
End effector f(6) & Closed ball  & $y_{1} (70,90,130)$ &  & $r_{1} = 16mm$ \\
\hline
Link g(0) & Capsule  & $a_{2} (0,120,0)$ & $b_{2} (0,120,65)$ & $r_{2} = 15mm$ \\
Link g(1) & Capsule  & $b_{2} (0,120,65)$ & $c_{2} (0,120,130)$ & $r_{2} = 15mm$ \\
Link g(2) & Capsule  & $c_{2} (0,120,130)$ & $d_{2} (35,120,130)$ & $r_{2} = 15mm$ \\
Link g(3) & Capsule  & $d_{2 }(35,120,130)$ & $e_{2} (70,120,130)$ & $r_{2} = 15mm$ \\
Link g(4) & Capsule  & $e_{2} (70,120,130)$ & $x_{2} (70,100,130)$ & $r_{2} = 15mm$ \\
Link g(5) & Capsule  & $x_{2} (70,100,130)$ & $y_{2} (70,30,130)$ & $r_{2} = 15mm$ \\
End effector g(6) & Closed ball  & $y_{2} (70,30,130)$ &  & $r_{1} = 16mm$ \\
\hline
\end{tabular}
\end{minipage}
\end{center}
\end{table}

In Table \ref{tab3}, $a_{1}$, $b_{1}$, $c_{1}$, $d_{1}$, $e_{1}$, $x_{1}$, and $y_{1}$ represent positional parameters of six capsules and a closed ball in $Robot$ 1.
$a_{2}$, $b_{2}$, $c_{2}$, $d_{2}$, $e_{2}$, $x_{2}$, and $y_{2}$ represent positional parameters of six capsules and a closed ball in $Robot$ 2.
$r_{1}$ is the radius of closed balls representing the end effectors of $Robot$ 1 and $Robot$ 2.
$r_{2}$ is the radius of capsules that represent the links of $Robot$ 1 and $Robot$ 2.
Then, based on the positional parameters in Table \ref{tab3}, we verify the collision of $Robot$ 1 and $Robot$ 2 in Theorem \ref{tworobotco}.
\\
\\
\begin{myThm}{Collision of two robots with positional parameters}
\label{tworobotco}
\end{myThm}
\begin{lstlisting}
&$\vdash$& !f g.
 f(0) = capsule_CGA vector[&\&&0;&\&&0;&\&&0]
                    vector[&\&&0;&\&&0;&\&&65] &\&&15 /\
 f(1) = capsule_CGA vector[&\&&0;&\&&0;&\&&65]
                    vector[&\&&0;&\&&0;&\&&130] &\&&15 /\
 f(2) = capsule_CGA vector[&\&&0;&\&&0;&\&&130]
                    vector[&\&&35;&\&&0;&\&&130] &\&&15 /\
 f(3) = capsule_CGA vector[&\&&35;&\&&0;&\&&130]
                    vector[&\&&70;&\&&0;&\&&130] &\&&15 /\
 f(4) = capsule_CGA vector[&\&&70;&\&&0;&\&&130]
                    vector[&\&&70;&\&&45;&\&&130] &\&&15 /\
 f(5) = capsule_CGA vector[&\&&70;&\&&45;&\&&130]
                    vector[&\&&70;&\&&90;&\&&130] &\&&15 /\
 f(6) = cball_CGA vector[&\&&70;&\&&90;&\&&130] &\&&16 /\
 g(0) = capsule_CGA vector[&\&&0;&\&&120;&\&&0]
                    vector[&\&&0;&\&&120;&\&&65] &\&&15 /\
 g(1) = capsule_CGA vector[&\&&0;&\&&120;&\&&65]
                    vector[&\&&0;&\&&120;&\&&130] &\&&15 /\
 g(2) = capsule_CGA vector[&\&&0;&\&&120;&\&&130]
                    vector[&\&&35;&\&&120;&\&&130] &\&&15 /\
 g(3) = capsule_CGA vector[&\&&35;&\&&120;&\&&130]
                    vector[&\&&70;&\&&120;&\&&130] &\&&15 /\
 g(4) = capsule_CGA vector[&\&&70;&\&&120;&\&&130]
                    vector[&\&&70;&\&&100;&\&&130] &\&&15 /\
 g(5) = capsule_CGA vector[&\&&70;&\&&100;&\&&130]
                    vector[&\&&70;&\&&30;&\&&130] &\&&15 /\
 g(6) = cball_CGA vector[&\&&70;&\&&30;&\&&130] &\&&16
 ==> collision (6,f) (6,g)
\end{lstlisting}
In Theorem \ref{tworobotco}, $\texttt {vector[\&0;\&0;\&0]}$ denotes a three-dimensional vector with parameter (0,0,0).
According to Definition \ref{rocomo}, $\texttt {collision (6,f) (6,g)}$ is equivalent to $\texttt {$\sim$(robot (0..6)}$ $\texttt {f INTER robot (0..6) g = \{\})}$.
We use the equivalent collision conditions of two single-arm industrial cooperative robots, shown in Theorem \ref{robotcoco}, to discover whether the components of two single-arm industrial cooperative robots collide or not.
During the verification process, we find that the collision occurs with the components represented by $\texttt {f(6)}$ and $\texttt {g(5)}$.
Combining with the automatic proof tactic in HOL Light, Theorem \ref{tworobotco} is verified by using the shortest distance models among different components of the robot.

The effort spent in verification details for each theorem in Sect.\ref{sec3} and Sect.\ref{sec4} is represented in the form of proof lines, the man-hours and the complexity analysis, shown as Table \ref{tab4}.
\begin{table}[h]
\begin{center}
\begin{minipage}{430pt}
\caption{Verification Details for Each Theorem}\label{tab4}%
\setlength{\tabcolsep}{1.5mm}
\begin{tabular}{|@{}|p{10cm}|p{1cm}|p{1cm}|p{1.75cm}|@{}|}
\hline
Formalized Theorems & Proof Lines & Man-Hours & Complexity Analysis\\
\hline
Theorem \ref{ballbou} Boundness of the closed ball & 90 & 10.5 & Medium \\ \hline
Theorem \ref{cenbou} Boundness of the center line & 22 & 2.5  & Easy  \\ \hline
Theorem \ref{cencoin} Equality of center line   & 10 & 1  & Easy\\  \hline
Theorem \ref{capcoin} Equality of the capsule  & 26 & 2.5  & Easy\\  \hline
Theorem \ref{mindiabc} Minimum value under different conditions   & 215 & 75  & Hard\\ \hline
Theorem \ref{theopse3} Minimum value if $ac - {b^2} \ne 0$ and $s_{1}{_k} , s_{2}{_k} \in [0,1]$   & 498 & 160  & Hard\\ \hline
Theorem \ref{disL1} Minimum value in Case 1  & 29 & 2.5  & Easy\\ \hline
Theorem \ref{babaco} Collision of two closed balls  & 52 & 6  & Medium\\ \hline
Theorem \ref{bacaco} Collision of the closed ball and the capsule  & 189 & 71  & Hard\\ \hline
Theorem \ref{cacaco} Collision of two capsules  & 104 & 67  & Hard \\ \hline
Theorem \ref{robotexi} The configuration rules of the robots  & 286 & 95  & Hard \\ \hline
Theorem \ref{sincoro} The single component of the robot  & 281 & 52  & Hard \\ \hline
Theorem \ref{equco} Equivalent conditions of two robots  & 62 & 7.5  & Medium \\ \hline
Theorem \ref{robotcoco} Equivalent collision conditions of two robots  & 43 & 5.5  & Medium \\ \hline
Theorem \ref{tworobotnoco} Collision of two arms of the service robot  & 448 & 154  & Hard \\ \hline
Theorem \ref{tworobotco} Collision of two arms with positional parameters  & 525 & 170  & Hard \\
\hline
\end{tabular}
\end{minipage}
\end{center}
\end{table}

The man-hours in Table \ref{tab4} are affected by two factors.
The first factor is the number of lines of HOL light code per hour by a person with average expertise.
The second factor is the complexity of the proved theorem.
Sometimes, to prove a more complex theorem, we need to provide some auxiliary theorems to verify the correctness of this theorem.
Meanwhile, simple arithmetic reasoning on vectors and real numbers is also required.
During the process of verification, we use the existing definitions, theorems and strategies in the HOL Light theorem library to prove our goal.

After the above analysis, we apply the formal verification framework for the robot collision detection method to the collision detection between two single-arm industrial cooperative robots.
Based on the shortest distance models of the geometric primitives, the collision models of the geometric primitives, and the formal model of the robot body, the general collision model and the special collision model with the given parameters of two single-arm industrial cooperative robots are constructed.
Therefore, the application of formal verification framework for the robot collision detection method to the collision detection between two single-arm industrial cooperative robots can clearly verifies the flexibility and reliability of the proposed framework.

\section{Conclusion}\label{sec5}

In this paper, we proposed a formal verification framework for the robot collision detection method in HOL Light, which is based on CGA.
Firstly, based on the CGA library, we formalized the closed ball, the center line and the capsule, the geometric primitives of the robot simplified geometric model, and formally verified their related properties.
Secondly, we established the shortest distance models of geometric primitives, which include the shortest distance model of two closed balls, the shortest distance model of the closed ball and the capsule, and the shortest distance model of two capsules.
Then, we verified the collision conditions of two closed balls, the closed ball and the capsule, and two capsules.
Finally, we realized the higher-order logic expression of the robot body and verified the equivalent conditions of the robot collision model.
To illustrate the complexity and flexibility of the formal verification framework of the robot collision detection method, we apply it to the collision detection of two single-arm industrial cooperative robots.

Overall our formalization requires more than 4000 lines of codes.
The main difficulties encountered are the enormous amount of user intervention required due to the uncertainty of the higher-order logic.
Further research should be directed to providing a formal verification framework for dynamic robot collision detection, which is essential to make them potentially usable in real-world applications.
The formally proven dynamic framework may realize that the cooperative robots can not collide because the cooperative robot slowed down soon enough to prevent the collision.

\section*{Acknowledgment}
This work was supported by National Natural Science Foundation of China (62272323, 62272322, 62002246), the National Key R \& D Plan (2019YFB1309900), the Project of Beijing Municipal Education Commission (KM201910028005, KM202010028010) and Academy for Multidisciplinary Studies, Capital Normal University.

\bibliographystyle{alphaurl}
\bibliography{myBibLib.bib}

\end{document}